# Scalable and interpretable rule-based link prediction for large heterogeneous knowledge graphs

Simon Ott, Laura Graf, Asan Agibetov, Christian Meilicke, Matthias Samwald

*Abstract*—Neural embedding-based machine learning models have shown promise for predicting novel links in biomedical knowledge graphs. Unfortunately, their practical utility is diminished by their lack of interpretability. Recently, the fully interpretable, rule-based algorithm AnyBURL yielded highly competitive results on many general-purpose link prediction benchmarks. However, its applicability to large-scale prediction tasks on complex biomedical knowledge bases is limited by long inference times and difficulties with aggregating predictions made by multiple rules.

We improve upon AnyBURL by introducing the SAFRAN rule application framework which aggregates rules through a scalable clustering algorithm. SAFRAN yields new state-of-the-art results for fully interpretable link prediction on the established general-purpose benchmark FB15K-237 and the large-scale biomedical benchmark OpenBioLink. Furthermore, it exceeds the results of multiple established embedding-based algorithms on FB15K-237 and narrows the gap between rule-based and embedding-based algorithms on OpenBioLink. We also show that SAFRAN increases inference speeds by up to two orders of magnitude.

## Motivation

Machine learning methods based on neural embedding models showed promising results for link prediction on large biomedical knowledge graphs [1]–[3]. Recently, the novel rule-based algorithm AnyBURL [4], [5] was shown to be competitive with neural embedding models on established general-domain link prediction benchmarks like FB15K-237 [6], while being much faster to train and being fully interpretable [5], [7]. In contrast to neural models, rule-based methods are fully transparent and predictions are intuitively explainable. These properties are highly valuable for link prediction in the biomedical domain, where both scalability and explainability are of importance.

However, AnyBURL currently has limitations: 1) while training is fast, prediction with large rule sets is often slow, 2) generated rules are often functionally redundant, making it difficult to meaningfully aggregate predictions made by different rules. To address these limitations, we introduce the SAFRAN ('Scalable and fast non-redundant rule application') software package and evaluate it on an established general-domain benchmark as well as a large-scale biomedical benchmark.

### A. Rule aggregation

Rule-based link prediction methods produce many rules with different confidences that are subsequently applied to the knowledge graph in order to generate candidates for a link prediction task. AnyBURL generates rules of three categories, which are explained in detail in [4]:

TABLE I
OVERVIEW OF POSSIBLE RULE TYPES GENERATED BY ANYBURL

| | |
|---|---|
| Cyclic (C) | $h(Y, X) \leftarrow b_1(X, A_2), \ldots, b_n(A_n, Y)$ |
| Acyclic 1 (AC1) | $h(c_0, X) \leftarrow b_1(X, A_2), \ldots, b_n(A_n, c_{n+1})$ |
| Acyclic 2 (AC2) | $h(c_0, X) \leftarrow b_1(X, A_2), \ldots, b_n(A_n, A_{n+1})$ |

X and Y are variables that appear in the rule head and body, $A_i$ is a variable that appears only in the rule body, $c_i$ is a constant.

Results of rules qualified for a prediction must be aggregated, as multiple rules can propose the same entity. This is commonly done by sorting proposed candidates by their highest confidence (max aggregation). If the confidences of multiple entities are the same, they are sorted by their second-best

This work was supported by European Community's Horizon 2020 Programme grant number 668353 (U-PGx).

S. Ott, L. Graf, A. Agibetov and M. Samwald are with the Section for Artificial Intelligence and Decision Support; Medical University of Vienna, Währinger Straße 25a, 1090 Vienna, Austria (e-mail: firstname.lastname@meduniwien.ac.at)

C. Meilicke is a member of the Data and Web Science Group; University Mannheim, B6, 26, 68159 Mannheim, Germany (e-mail: christian@informatik.uni-mannheim.de)

confidence (i.e. the second-best confidence of the rule that produced the entity).

Unfortunately, max aggregation is constrained to relatively simple predictions: Each prediction is only informed by a single rule with the highest confidence; it is not possible to make predictions based on a weighted combination of different rules. This limitation is potentially addressed by noisy-or aggregation. Instead of taking the maximum of confidences, the probability that at least one of the rules proposed the correct candidate is used for generating a list of ranked candidates. This approach assumes independence / non-redundancy of all rules. However, most rules are in some way redundant with other rules, and noisy-or was found to perform worse than the max aggregation in evaluations because of this [4].

### B. Direct and indirect rule redundancy

Let r be a rule, then a substitution $\theta$ is a function that maps r to a rule $\theta(r)$ within the language bias of AnyBURL where each occurence of a constant is replaced by a variable. Two rules $r_1$ and $r_2$ are directly redundant if there is a substitution $\theta_1$ and $\theta_2$ such that $\theta_1(r_1) = \theta_2(r_2)$. This definition ensures that directly redundant rules $r_1$ and $r_2$ create the same candidate for a given completion task only if their groundings are the same, i.e., only if they fire for the same reason. Table 2 provides basic and intuitive examples of directly redundant rules. (1) and (2) are directly redundant, as there exists an equivalent substitution $\theta$ for both rules and they generate the same candidates for the same reason when X = serotonin. As D-Amino Acid Oxidase (DAO) oxidises D-amino acids such as D-alanine [8] there is also direct redundancy between (3) and (4), because both rules answer the completion task (DAO, gene_drug, ?), inter alia, with the entity alanine.

TABLE II
EXAMPLES OF DIRECTLY REDUNDANT RULES: (1) AND (2), (3) AND (4)

| drug_reaction_gene(X,Y) | <= | drug_activation_gene(X,Y) |
|---|---|---|
| drug_reaction_gene(serotonin,Y) | <= | drug_activation_gene(serotonin,Y) |
| gene_drug(X,D-alanine) | <= | drug_catalysis_gene(D-alanine,X) |
| gene_drug(DAO,Y) | <= | drug_catalysis_gene(Y,DAO) |

Object Identity [9], [10] refers to a constraint on the variables of a rule, which states that no instantiations of two distinct variables or the instantiation of a variable and the constant of a rule can be the same. Since AC2 rules are therefore constrained by $c_0 \neq A_{n+1}$, Object Identity prevents a direct redundancy between rules of type C and AC2, as well as AC1 (if $c_0 = c_{n+1}$) and AC2.

Indirect redundancy occurs between two rules $r_1$ and $r_2$ where $\theta_1(r_1) \neq \theta_2(r_2)$, which infer entities for the same reason (i.e. express the same knowledge). As an example, Rule 5 and 6 both encompass the same drug-gene interaction. The enzyme CYP2D6, which metabolizes about a quarter of all prescribed drugs, is primarily expressed in the liver and has a high phenotypic variability [11]. Individuals who are ultrarapid metabolizers (i.e. CYP2D6 is overexpressed), metabolize e.g. codeine into morphine more rapidly, which can lead to significant overdose effects such as dyspnea. As another example, rule 7 is used for prediction of drugs that cause fatigue/tiredness based on whether the drug is an antagonist of the histamine receptor HRH1. One such drug could be Mirtazapine [12], a tetracyclic antidepressant with a high binding affinity to HRH1. Rule 8 entails rule 7, as it predicts such drugs, based on the ability of the drug to bind and inhibit a protein that is part of the histamine receptor binding pathway. Both examples could lead to an overestimation of the confidence of an entity. Valid entities that are only proposed by one rule on the other hand, e.g. due to incomplete knowledge, could be ranked too low.

## II. METHODS

We introduce SAFRAN, a scalable rule inference algorithm which addresses rule redundancy by clustering dependent rules prior to aggregating results using the noisy-or approach [1]. Similarity of two rules is measured by the similarities of the sets of entities that can be inferred by them. This set, which we shall call the *solution set* of a rule, contains all possible head-tail pairs of the head atom of a rule that can be grounded by the rule in the training set. Note that if the rule is of type AC1 or AC2 either the head or the tail of all pairs in its solution set is fixed. In this work the similarity of two sets is expressed by the Jaccard coefficient, the size of the intersection of both sets divided by the size of the union. As the calculation of the Jaccard coefficient is very inefficient for large sets, the Jaccard coefficient is estimated using the MinHash scheme [13], which makes time complexity linear and memory usage constant. Clusters of rules are formed by assigning rules above a certain Jaccard index to the same cluster using depth first search (DFS) on a weighted dependency graph, where the nodes are rules and the weight of the edge between each two rules represents the Jaccard index. The algorithm can only traverse along an edge if the Jaccard is greater than the threshold. Furthermore, the maximum confidences of entities proposed by each cluster are joined using Noisy-Or. Figure 1 shows the 6 distinct threshold parameters of and between the rule types. SAFRAN uses one of two search strategies for finding the optimal thresholds: grid search (parameter sweep) and random search. For each iteration of the random search the 6 thresholds are randomly sampled. For grid search, the range of the Jaccard index (0.0-1.0) is divided into $n$ equally distant thresholds and each threshold is subsequently used for clustering. As performing a grid search for 6 parameters would require $n^6$ iterations, only one threshold is used for all parameters with this strategy, which only requires $n$ iterations.

TABLE III
EXAMPLES OF INDIRECTLY REDUNDANT RULES: (5) AND (6), (7) AND (8)

| drug_phenotype(X, dyspnea) | <= | drug_binding_gene(X, A), gene_overexpressed_anatomy(A, liver) | (5) |
|---|---|---|---|
| drug_phenotype(X, dyspnea) | <= | drug_binding_gene(X, A), gene_phenotype(A, abnormality_of_metabolism) | (6) |
| drug_phenotype(X, fatigue) | <= | drug_bindinhibition_gene(X, hrh1) | (7) |
| drug_phenotype(X, fatigue) | <= | drug_bindinhibition_gene(X, A), gene_pathway(A, histamine_receptor_binding) | (8) |

---

[1] https://github.com/OpenBioLink/SAFRAN

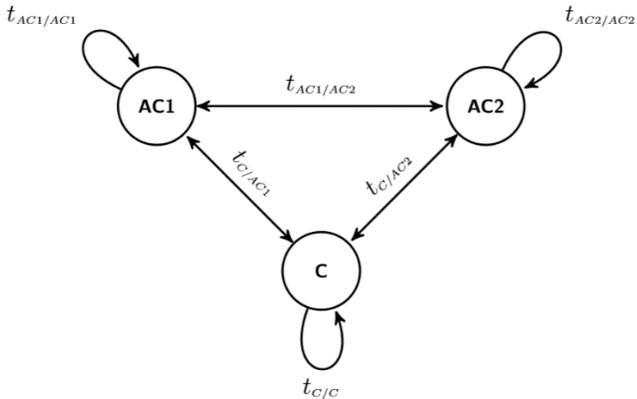

Fig. 1.  Threshold parameters of and between rule types

The fitness of the clusters for predicting new links is evaluated on the validation set using the mean-reciprocal rank based on the top k predicted entities. The mean-reciprocal rank is calculated according to the BOTTOM evaluation protocol for dealing with same score entities [14], where the correct entity is inserted at the end of a list of such entities. The rationale behind this ranking is to prevent the algorithm taking advantage of the underlying data structure for finding optimal cluster assignments. An extreme example of such a misuse would be to answer all knowledge tasks with all possible entities and the same confidence, resulting in some hits by chance.

To achieve scalability on large knowledge bases, a compiled low-level programming language (C++) was selected for the implementation of SAFRAN. The adjacency matrices of all relations of a graph are stored in compressed sparse row format, which enables fast and memory efficient queries of neighborhoods. Avoiding unnecessary inferences of the same rule and several fine-tuned optimizations of code parts that are executed frequently further improved memory usage and runtime.

## A. Experimental Setup

We evaluated SAFRAN on the general-domain FB15K-237 benchmark and the biomedical OpenBioLink [15] benchmark. We compared the performance of SAFRAN with existing AnyBURL inference algorithms, other rule based ('white box') as well as embedding based ('black box') state-of-the-art link prediction methodologies. Training, inferences and evaluations were run on a machine with 24 physical (48 logical) Intel(R) Xeon(R) CPU E5-2650 v4 @ 2.20GHz cores, 264 GB of RAM and 10 Nvidia GTX1080Ti GPUs. GPU-accelerated computing was only used for training embedding-based methods; SAFRAN and AnyBURL only utilized CPUs. For optimal comparison with the best results using the maximum aggregation approach presented in [5], rulesets used for the results were learned with AnyBURL for 1000 seconds each using 22 threads. The same rulesets were used for inference and evaluation with AnyBURL and SAFRAN. Embedding-based models were trained and evaluated with the DGL-KE software package [16]; the best results of extensive hyperparameter searches for embedding-based models are reported.

## III. RESULTS

Table 4 summarizes the results of applying SAFRAN and other methods to link prediction benchmarks. SAFRAN outperforms AnyBURL and all other white box link prediction algorithms on both benchmarks and all measures. The gains of SAFRAN compared to AnyBURL were greater for the large-scale, complex biomedical data captured by the OpenBioLink benchmark. The random search variant of SAFRAN produced better results than the parameter sweep variant. SAFRAN reduced the gap to the predictive performance of black box, embedding-based methods, but was still outperformed by some of the black box models.

TABLE IV
RESULTS OF APPLYING DIFFERENT RULE AGGREGATION METHODS ON RULES LEARNED FROM THE FB15K-237 AND OPENBIOLINK BENCHMARK DATASETS. BEST EXPLAINABLE / WHITE-BOX RESULT IS DENOTED IN BOLD, BEST OVERALL RESULT (INCLUDING BLACK-BOX) IS UNDERLINED.

| | | Hits@1 | Hits@3 | Hits@10 |
|---|---|---|---|---|
| **FB15k-237** | | | | |
| AnyBURL Maximum | | 0.2737 | 0.3884 | 0.5228 |
| AnyBURL Noisy-Or | | 0.2228 | 0.3298 | 0.4621 |
| SAFRAN Non-redundant Noisy-Or [a] | grid search (single threshold, resolution = 0.005) | 0.2888 | 0.4046 | 0.5346 |
| | random search (multi threshold, resolution = 0.1, iterations = 10000) | **0.3013** | **0.4175** | **0.5465** |
| *Comparison: Symbolic / White box* | | | | |
| AMIE+ [b] | | 0.174 | | 0.409 |
| RuleN [b] | | 0.182 | | 0.42 |
| RLvLR [b] | | | | 0.393 |
| Neural LP [c] | | 0.166 | 0.248 | 0.348 |
| MINERVA [c] | | 0.217 | 0.329 | 0.456 |
| *Comparison: Embedding / Black box* | | | | |
| RESCAL [d] | | 0.263 | | 0.541 |
| TransE [d] | | 0.221 | | 0.497 |
| DistMult [d] | | 0.250 | | 0.531 |
| ComplEx [d] | | 0.253 | | 0.534 |
| ConvE [d] | | 0.248 | | 0.521 |
| **OpenBioLink** | | | | |
| AnyBURL Maximum | | 0.1948 | 0.3066 | 0.4630 |
| AnyBURL Noisy-Or | | 0.0754 | 0.1513 | 0.4217 |
| SAFRAN Non-redundant Noisy-Or [a] | grid search (single threshold, resolution = 0.005) | 0.2205 | 0.3424 | 0.5056 |
| | random search (multi threshold, resolution = 0.1, iterations = 10000) | **0.2232** | **0.3473** | **0.5110** |
| *Comparison: Embedding / Black box* | | | | |
| RESCAL | | <u>0.407</u> | <u>0.479</u> | <u>0.615</u> |
| TransR | | 0.369 | 0.451 | 0.592 |
| DistMult | | 0.184 | 0.331 | 0.534 |
| ComplEx | | 0.166 | 0.314 | 0.525 |
| RotatE | | 0.156 | 0.315 | 0.522 |
| TransE | | 0.128 | 0.268 | 0.441 |

[a] our approach, [b] reported in [5], [c] reported in [17], [d] reported in [18], while all other results were generated within this study.

Table 5 compares runtimes of SAFRAN with AnyBURL for the rulesets and datasets used in Table 4. A major bottleneck in rule application is the inference and filtering of entities proposed by different rules. Noisy-Or requires the inference and filtering of all rules, where SAFRAN achieves an estimated speedup by a factor of 790.2 on the OpenBioLink ruleset compared to the AnyBURL inference engine. Maximum aggregation takes less time as it does not require the inference of all rules. In the maximum aggregation setting, SAFRAN achieved a speedup of 50.1.

TABLE V
COMPARISON OF ANYBURL AND SAFRAN RUNTIMES (22 THREADS, INTEL(R) XEON(R) CPU E5-2650 V4 @ 2.20GHZ). NOTE THAT MAXIMUM AGGREGATION TAKES LESS TIME THAN THE INFERENCE OF THE COMPLETE RULESET, AS MAXIMUM AGGREGATION, UNLIKE NOISY-OR, DOES NOT REQUIRE THE INFERENCE OF ALL RULES.

|  | AnyBURL | SAFRAN | Speed up |
|---|---|---|---|
| **FB15k-237** | | | |
| Rule inference (inference of complete ruleset) | 615 259 ms | 46 205 ms | 13.3 x |
| Maximum aggregation | 24 523 ms | 4222 ms | 5.8 x |
| **OpenBioLink** | | | |
| Rule inference (inference of complete ruleset) | 2252 h [a] | 2.85 h | 790.2 x |
| Maximum aggregation | 35.9 h | 43 min | 50.1 x |

[a] Estimation: After a runtime of 12 hours all rules were inferenced for 975 prediction tasks.

## IV. DISCUSSION AND FUTURE WORK

We introduced SAFRAN, a rule inference engine that can be applied to rulesets generated by AnyBURL and that achieves new state-of-the-art results for white box, fully interpretable link prediction on the widely established FB15k-237 benchmark as well as the large-scale biomedical benchmark OpenBioLink. While some black box, embedding-based methods still outperform rule based methods, there are many practical advantages of white box methods that, in practice, might make them preferable in biomedical applications, such as increased trustworthiness, ability to ascertain the soundness of learned models and greater utility for hypothesis generation.

Although the introduced algorithm yields better predictions than maximum aggregation or simple noisy-or aggregation, more research should be devoted to the creation of better clusters. The Jaccard index is a good indicator of the redundancy of two rules, but – since knowledge graphs are incomplete – not a perfect one. Searching randomly for more predictive clusters sometimes outperforms our methodology for certain relations. Iterating over all rules and discarding rules that lower the performance of the noisy-or approach works, but fails to find the global optimum, as only immediate reward is consumed. Reinforcement learning methods like temporal-difference learning or metaheuristics, e.g. genetic algorithms, could further improve the predictiveness of the noisy-or approach.

Rule learning itself could profit from further information about communities of nodes in the graph, which could be added to the graph in the form of auxiliary entities and relations in a pre-processing step. This would enable rule learners like AnyBURL to detect patterns that are difficult to express as rules that are solely based on relationships explicitly found in the knowledge graph, and could further improve predictive accuracy.